\documentclass{esannV2}
\usepackage[dvips]{graphicx}
\usepackage[latin1]{inputenc}
\usepackage{amssymb,amsmath,array}
\usepackage{float}
\usepackage[ruled, linesnumbered]{algorithm2e}  
\usepackage{mathtools}

\usepackage{multirow}
\usepackage{hhline}
\usepackage{hyperref}
\hypersetup{hidelinks}

\newcommand{\res}[2]{$#1\,{\scriptstyle \pm\, #2}$}
\newcommand{\bres}[2]{$\textbf{#1}\,{\scriptstyle \pm\, \textbf{#2}}$}
\usepackage{common}

\begin{document}
\title{Federated Adaptation of Reservoirs via Intrinsic Plasticity}

\author{Valerio De Caro, Claudio Gallicchio and Davide Bacciu
\vspace{.3cm}\\
University of Pisa - Department of Computer Science\\
Largo Bruno Pontecorvo, 3, 56127, Pisa - Italy
}
\maketitle

\begin{abstract}
We propose a novel algorithm for performing federated learning with Echo State Networks (ESNs) in a client-server scenario. In particular, our proposal focuses on the adaptation of reservoirs by combining Intrinsic Plasticity with Federated Averaging. The former is a gradient-based method for adapting the reservoir's non-linearity in a local and unsupervised manner, while the latter provides the framework for learning in the federated scenario. We evaluate our approach on real-world datasets from human monitoring, in comparison with the previous approach for federated ESNs existing in literature. Results show that adapting the reservoir with our algorithm provides a significant improvement on the performance of the global model.
\end{abstract}
\section{Introduction}
On-the-edge applications are requiring an increasingly frequent use of machine learning (ML) systems, a key component for the success of human-centric cyber-physical systems \cite{bacciu2021teaching,de2022ai}. This type of applications involves large numbers of participating users, each of whom is a source of inherently sequential data. This setting involves two main challenges: (1) the devices involved in this domain are often low-powered, and achieving good trade-off between performance in tasks on temporal data and efficiency may be difficult; (2) the data may be subject to privacy constraints, which does not allow the use of common distributed learning techniques. The former is approached through the use of Echo State Networks \cite{jaeger2001echo}, a recurrent neural network characterized by excellent accuracy-efficiency trade-off. The latter has been tackled with Federated Learning \cite{mcmahan2017communicationefficient}, a distributed learning method in which a global model is learned without transferring local participant's data. Notwithstanding, to the best of our knowledge, the intersection area of these two approaches is limited to \texttt{IncFed} \cite{bacciu2021federated}, a method for performing an exact computation of the readout of ESNs in a federated scenario.

In this paper, we aim to further bridge the gap between these two areas by proposing Federated Intrinsic Plasticity (\texttt{FedIP}), a novel algorithm for performing the unsupervised adaptation of a federation of reservoirs. Our proposal is based on Intrinsic Plasticity \cite{schrauwen2008improving}, an existing algorithm for adapting the dynamics of a reservoir with respect to the input sequence, and Federated Averaging, a client-server algorithm for learning a global model by averaging models learned from local client data. We assess the algorithm on two Human Activity Recognition benchmarks, and show that, with a low computation and communication overhead from the use of \texttt{FedIP}, the performance of the learned global model can improve significantly.
\section{Federated Adaptation of Reservoirs}
\paragraph{Echo State Networks.} Reservoir Computing (RC) \cite{lukosevicius2009reservoir} is a paradigm which leverages on the evolution of the neural activations of Recurrent Neural Networks (RNNs) as a discrete-time dynamical system. A remarkable example of this paradigm is represented by Echo State Networks (ESNs) \cite{jaeger2001echo}, which allow to perform learning task on sequential data efficiently. ESNs are made up of two main components: a \textit{reservoir}, a recurrent layer of sparsely connected neurons, holding the internal state which evolves over time; a \textit{readout}, a linear transformation on the domain of the reservoir states. Formally, given an input sequence of vectors $\vu(t) \in \Real^{N_U}$ with $t \in [T]$, the equations modelling the state transition of the reservoir with leaky-integrator neurons and the transformation applied by the readout can be described as
\begin{align}\label{eq:li-esn}
\begin{split}
    \vx(t) = &\;(1-a)\,\vx(t-1) + a\tanh\left(\textbf{W}_{in}\vu(t)+\vb_{rec}+\hat{\mW}\vx(t-1)\right)\\
    \vy(t) = &\;\textbf{W}\vx(t) + \vb_{out}
\end{split}
\end{align}
where $\mW_{in} \in \Real^{N_R \times N_U}$ is the input-to-reservoir weight matrix, $\hat{\mW} \in \Real^{N_R \times N_R}$ is the recurrent reservoir-to-reservoir weight matrix, $\vb_{rec} \in \Real^{N_R}$ is the reservoir bias term, $\mW \in \Real^{N_Y \times N_R}$ is the readout weight matrix, $\vb_{out}$ is the output bias term, $a \in (0, 1]$ is the leaking rate and $\vx(0) = \textbf{0}$. Instead of backpropagating the error signal through time as in standard RNNs, ESNs keep the input-to-reservoir matrix $\mW_{in}$ and the reservoir-to-reservoir matrix $\hat{\mW}$ \textbf{fixed}, with the only constraint of choosing spectral radius $\rho(\hat{\mW}) < 1$ to ensure the stability of the dynamical system. Thus, only the readout weights $\mW$ are learned. This allows to solve the optimization problem as a linear system with the closed form $\mW = \textbf{Y}\textbf{S}^T(\textbf{S}\textbf{S}^T + \lambda\textbf{I})^{-1}$, where $\textbf{Y}$ denotes the set of target labels, $\textbf{S}$ is the set of reservoir states, $\lambda$ is an L2-regularization term, and $\textbf{I}$ is the identity matrix. In \cite{bacciu2021federated}, a federated version of this form was proposed, where, given the local matrices of clients $\textbf{A}_c = \textbf{Y}_c\textbf{S}_c^T$ and $\textbf{B}_c = \textbf{S}_c\textbf{S}_c^T$, the server is able to compute the exact solution of the linear system while complying to the privacy constraints of the federated setting.

\paragraph{Adapting Reservoirs.} Intrinsic Plasticity (IP) \cite{schrauwen2008improving} is an algorithm inspired by a biological phenomenon, called \textit{homeostatic plasticity}, for adapting the reservoir in an unsupervised manner. Focusing the attention on a single neural unit, the algorithms requires the neuron's function to be reformulated as $\Tilde{x} = \tanh{(gx_{net}+b)}$, where $g$ and $b$ are the gain and the bias of the non-linearity, respectively, and $x_{net}$ is the net input to the neuron. When using the $\tanh$ as non-linearity, the objective of IP is to minimize the Kullback-Leibler divergence between the empirical distribution of the neuron's activations and a desired Gaussian distributions with parameters $\mu$ and $\sigma$. This is performed with a gradient-based approach in which the update rules are formalized as
\begin{align}
    \label{eq:deltab}&\Delta b = -\eta ((-\mu / \sigma^2) + (\Tilde{x} / \sigma^2 + 1 - \Tilde{x}^2 + \mu \Tilde{x})) \\
    \label{eq:deltag}&\Delta g = \eta / g + \Delta b x_{net}
\end{align}
where $\mu$ and $\sigma$ denote the mean and standard deviation of the target Gaussian distribution and $\eta$ is a learning rate. This simple learning rules allow to maximize the information content of reservoir states, and to reduce the variance in performance caused by bad initializations.

\paragraph{Federated Intrinsic Plasticity.} In this paper, we extend the use of IP to a federated scenario. Our proposal is intended for a client-server topology, and is based on the Federated Averaging (\texttt{FedAvg}) \cite{mcmahan2017communicationefficient} algorithm. Given a server $S$, a set of clients $[C]$ and a number of local epochs $E$, a generic round $t$ of \texttt{FedAvg} can be summarized by the following steps: (1) the server $S$ sends the global model parameters $\theta_t$ to all the clients; (2) starting from $\theta_t$, all the clients $c \in [C]$ perform $E$ epochs of adaptation on their local dataset; (3) all the clients $c \in [C]$ send their updated model $\theta_{t+1}^c$ to the server $S$; (4) the server $S$ applies the aggregation rule to the local models, producing $\theta_{t+1} = \sum_{c \in [C]} (n_c/n)\,\theta_{t+1}^c$, where $n_c$ is the cardinality of the local dataset of client $c$, and $n = \sum_{c \in [C]} n_c$. The weighting term $n_c/n$ in the aggregation rule allows to balance the update in the phase of statistical heterogeneity across clients. 

Our algorithm, namely \textbf{Federated Intrinsic Plasticity} (\texttt{FedIP}), instantiates \texttt{FedAvg} only on the gain and bias parameters, i.e. $\vg$ and $\vb$ (the pseudocode is summarized in Algorithm \ref{alg:fedip}). In particular, the server sends all the parameters of the reservoir (including $\vg$ and $\vb$) and the hyperparameters only in the initialization phase. Instead, in a generic round $t$, the only parameters updated and exchanged between clients and server are $\vg$ and $\vb$. In the step (3) a client $c$ performs $E$ iterations of IP on the local data and sends $\vg_{t+1}^c$ and $\vb_{t+1}^c$ to the server, while in (4) the server aggregates $\vg_{t+1}$ and $\vb_{t+1}$ as in \texttt{FedAvg}. performing the local adaptation of reservoirs via Intrinsic Plasticity, i.e. updating the parameters $\textbf{g}$. One major advantage of this algorithm is that, in each of communication round, the amount of parameters exchanged between client and server is rather small, which makes the algorithm suitable for on-the-edge scenarios.

\begin{algorithm}[t]
\label{alg:fedip}
  \SetKwInput{KwInput}{Input}
  \KwInput{clients $C$, learning rate $\eta$, local epochs $E$, batch size $B$}
  $\mathcal{R} \gets \{\mW_{in},\hat{\mW}, \mW\}$\;
  $\vg_0, \vb_0 \gets \textbf{1},\,\textbf{0}$\;
  Send $\mathcal{R}$, $\eta$, $T$ to all clients $c \in [C]$\;
  \For{{\upshape each round} $t \in \left\{1, 2, \dots\right\}$}{
    \For{{\upshape each client} $c \in [C]$ {\upshape \textbf{in parallel}}}{
      Send $\vg_t,\vb_t$ to client $c$\;
      $\vg_{t+1}^{c},\,\vb_{t+1}^{c} \gets $ $\text{LocalIPUpdate}_c$($\vg_t$, $\vb_t$, $\eta$, $E$) \tcp*{Eq. (\ref{eq:deltag}),(\ref{eq:deltab})}
    }
    $\vg_{t+1},\;\vb_{t+1}  \gets \sum_{c \in [C]} \frac{n_c}{n}\,\vg_{t+1}^c,\; \sum_{c \in [C]} \frac{n_c}{n} \vb_{t+1}^c$\;
  }
  \caption{Federated Intrinsic Plasticity (\texttt{FedIP})}
\end{algorithm}
\section{Experimental Assessment}
We compared the performance of the Echo State Networks in a federated setting with two algorithms: \texttt{IncFed}, proposed in \cite{bacciu2021federated}, and \texttt{FedIP} + \texttt{IncFed}. We tested both the algorithms on two well-known Human Activity Recognition benchmarks, i.e., WESAD \cite{schmidt2018wesad} and HHAR \cite{hhar}. The former is a dataset for stress and affect detection from wearable devices, which was collected from 15 participants in a $\sim$36-minute session where they performed activities depending cognitive state to be induced. The latter was collected from 9 users keeping 12 smart devices while performing different activities, with the aim of showing the heterogeneity of the sensing across the devices. Both the datasets lend themselves to adaptation to a federated scenario, since the data are equipped with a label which denotes the user they were gathered from. After performing a model selection for selecting the best hyperparameters of both the ESN and the corresponding algorithm, we tested the performance of the model on a subset of held-out users. This experiment was repeated with different percentages of users involved in the training phase.

\begin{table}
    \centering
    \caption{Hyperparameters tested on the two benchmarks. All the hyperparameters are common to both the algorithms, except for $\mu$, $\sigma$ and $\eta$ which are specific for \texttt{FedIP}.}
    \begin{tabular}{|c|c|c|c|}
            \cline{3-4}
            \multicolumn{2}{c|}{} & WESAD & HHAR\\\cline{1-4}
            
            \multirow{5}{*}{Both} & Units & \{200, 300, 400\} & \{100, 200, 300, 400, 500\}\\
            & $\rho(\hat{\mW})$ & $[0.3, 0.99)$ & $[0.3, 0.99)$ \\
            & Input scaling & $[0.5, 1)$ & $[0.5, 1)$ \\
            & Leakage $\alpha$ & $[0.1, 0.8]$ & $[0.1, 0.5]$ \\
            & L2 $\lambda$ & $[1e^{-4}, 1] $ & $[1e^{-4}, 1] $\\\hhline{|=|=|=|=|}
            
            \multirow{3}{*}{\texttt{FedIP}} & $\mu$ & 0 & 0 \\
            & $\sigma$ & $(0.005, 0.15)$ & $(0.005, 0.15)$ \\
            & $\eta$ & $0.01$ & $0.01$ \\
            & Epochs & $[3, 5, 10]$ & $[3, 5, 10]$ \\\hline

            \hline
        \end{tabular}
    \label{tab:hparams}
\end{table}
\paragraph{Setup} In WESAD, we used a subset of the available data, which consisted in 8 synchronized time-series of physiological data sampled at 700Hz by a chest-worn device. Each sample had a label corresponding to one of the 4 expected cognitive states of the user. In HHAR, we selected only the samples corresponding to the LG Nexus4 smartphone (which has a sample rate of $\sim$200Hz). Each sample had 6 features corresponding to the axes of the accelerometer and the gyroscope of the device, and a label denoting one of the 6 activities performed by the user. Then, we split both the dataset to user-specific chunks (15 for WESAD, 9 for HHAR), which simulate the local data of the clients in the federated setting. Each of the chunks was then normalized and split in non-overlapping sequences of 700 samples for WESAD (1 second) and 400 samples for HHAR ($\sim$2 seconds). For both datasets, we performed a training-validation-test split of the users. We used the validation split to monitor the performance of the models and rank them in the model selection, while the test split is used to assess the performance on clients joining the federation after the training is over. The splits are 9-3-3 and 5-2-2 for WESAD and HHAR respectively. For each experiment, first, we performed a random search with 30 trials to select the configuration that performed best on the given benchmark (the hyperparameter space is summarized by Table \ref{tab:hparams}). Then, we re-trained three instances of the selected configuration, and assessed the empirical risk on the test users.

Given the two benchmarks, we conducted the experiments with both \texttt{IncFed} and \texttt{FedIP}, and with different percentages of training users, to assess the performance of the algorithms with different availabilities of clients. We make the code publicly available to reproduce all the experiments\footnote{\href{https://github.com/vdecaro/federated-esn/}{https://github.com/vdecaro/federated-esn/}}.

\begin{table}
    \centering
    \caption{Results of the experiments on WESAD and HHAR datasets. For each percentage of the users, we report the mean and standard deviation of the test accuracy of each model.}
    \begin{tabular}{|c||c|c||c|c|}
            \hline
            \multirow{2}{*}{Training Users} & \multicolumn{2}{c||}{WESAD} & \multicolumn{2}{c|}{HHAR}\\\cline{2-5}
            & \texttt{IncFed} & \texttt{FedIP} & \texttt{IncFed} & \texttt{FedIP} \\\hline
            25\% & \res{75.24}{1.63} & \bres{75.45}{2.20} & \res{75.50}{2.79} & \bres{80.72}{0.99} \\
            50\% & \res{73.04}{0.70} & \bres{75.92}{0.84} & \res{77.06}{0.43} & \bres{82.24}{0.30} \\
            75\% & \res{75.64}{1.23} & \bres{81.15}{0.26} & \bres{80.67}{0.75} & \res{79.50}{1.01} \\
            100\%& \res{75.42}{0.78} & \bres{80.42}{0.79} & \res{81.47}{0.77} & \bres{82.09}{0.29} \\\hline
        \end{tabular}
    \label{tab:res}
\end{table}

\paragraph{Results} Table \ref{tab:res} shows the mean and standard deviation of the test accuracy over three retraining runs for each percentage of the training users on the two benchmarks. Both benchmarks show that using the federated version of Intrinsic Plasticity can benefit to the performance of the model on the task at hand. On the WESAD dataset, it is clear that \texttt{IncFed} suddenly meets an upper bound which is independent on the number of users involved in the training phase. This highlights that the dynamics of a reservoir initialized in a na\"{i}ve way does not allow to appropriately express features which are useful for discriminating the correct label. Instead, with the use of \texttt{FedIP}, the performance of the model grows with the number of users involved in the training phase, outperforming the performance of \texttt{IncFed} by 5 accuracy points when involving all the users in the training phase. In the HHAR benchmark, instead, we can observe an opposite trend from WESAD. While the performance of \texttt{IncFed} grows with the number of users involved in the training phase, \texttt{FedIP} allows to adapt the reservoir to dynamics which allow to generalize on the task even with the 25\% of users. However, the centralized version of IP is able to mitigate the variance of performance with respect to bad reservoir initializations, an effect that \texttt{FedIP} does not produce in our scenario. This requires further investigation to understand the effect of statistical heterogeneity among clients, a key determinant in federated learning scenarios. 

\section{Conclusions}
In this work we extended Intrinsic Plasticity (IP), a method which adapts the reservoir dynamics of ESNs with respect to the input sequence, to the federated setting. Our algorithm, namely \texttt{FedIP} and based on \texttt{FedAvg}, is intended for a client-server topology: in each round, clients perform $E$ epochs of local update via IP, and the server aggregates the updated gains and biases of the local model into the global one. Experiments showed that the gradient-based nature of IP suited the approach of \texttt{FedAvg}, and using \texttt{FedIP} for adapting the reservoirs in a distributed manner improved significantly the performance of the global model on both  benchmarks.

\bibliographystyle{unsrt}
{\footnotesize
\bibliography{references}}
\end{document}